# ROBUST AND REAL TIME DETECTION OF CURVY LANES (CURVES) WITH DESIRED SLOPES FOR DRIVING ASSISTANCE AND AUTONOMOUS VEHICLES

Amartansh Dubey[1] and K. M. Bhurchandi[2]

[1]Department of Electronics and Communication Engineering, Visvesvaraya Natioanal Institute of Technology, Nagpur, India
dubeyamartansh@gmail.com

[2] Department of Electronics and Communication Engineering, Visvesvaraya Natioanal Institute of Technology, Nagpur, India
bhurchandikm@ece.vnit.ac.in

## ABSTRACT

*One of the biggest reasons for road accidents is curvy lanes and blind turns. Even one of the biggest hurdles for new autonomous vehicles is to detect curvy lanes, multiple lanes and lanes with a lot of discontinuity and noise. This paper presents very efficient and advanced algorithm for detecting curves having desired slopes (especially for detecting curvy lanes in real time) and detection of curves (lanes) with a lot of noise, discontinuity and disturbances. Overall aim is to develop robust method for this task which is applicable even in adverse conditions. Even in some of most famous and useful libraries like OpenCV and Matlab, there is no function available for detecting curves having desired slopes , shapes, discontinuities. Only few predefined shapes like circle, ellipse, etc, can be detected using presently available functions. Proposed algorithm can not only detect curves with discontinuity, noise, desired slope but also it can perform shadow and illumination correction and detect/ differentiate between different curves.*

## 1. INTRODUCTION

Nowadays, one of the biggest topics under research is self-driving vehicles or smart vehicles which generally based on GPS synchronization, Image processing and stereo vision. But these projects are only successful on well-defined roads and areas and it is very hard to localize vehicles in real time when lanes or boundaries of roads are not well defined or have sharply varying slopes, discontinuities and noise. Even in some of most famous and useful libraries like OpenCV and Matlab, there is no function available for detecting curves of desired slopes, shapes, discontinuities. Only few predefined shapes like circle, ellipse, etc, can be detected using presently available functions. Using color thresholding techniques for detecting lanes and boundaries is very bad idea because it is not a robust method and will fails in case of discontinuity of lanes, shadow/illumination disturbances and noise present on roads.

For proper localization of a vehicle in between the desired region of interest (like between lanes or boundaries), it is very important to detect lanes and boundaries efficiently and it is also important to develop feedback algorithm for auto correction of vehicle motion in synchro with varying slopes of lanes or boundaries. It is also very important to avoid confusion due to multiple lanes or shadow/illumination disturbances and noises which have similar features as lanes or boundaries.

For solving all the above mentioned problems, we developed algorithm which is combination of four important algorithms, out of which two algorithms are new and unique and very efficient. Four algorithms are:

1) Dissection of curves into large number of Hough lines (called Curve stitching) and then implementing Mean value theorem using weighted centroids to keep track of curves which works well in adverse conditions like discontinuous curves (lanes).
2) Differentiating various curves (lanes) based on slope of curves. This helps in developing algorithm for tracking varying slopes of lane and develops efficient feedback system for proper localization in between roads.
3) Clustering of Hough lines to avoid confusion due to other lane type noises present on road.
4) Shadow, illumination correction and noise filters implemented on HSV color frame.
5) Feedback algorithm to continuously compare parameters (road texture) of current image frame with previous frame of image to decide boundaries in case of missing lanes or well defined boundaries.

All the above algorithms are described below one by one.

## 2. DISSECTION OF CURVE INTO INFINITE HOUGH LINES (CURVE STITCHING) TO IMPLEMENTING MEAN VALUE THEOREM USING WEIGHTED CENTROIDS.

Curve stitching is process of making curves and circles using straight lines. This means that a curve can be made using large number of tangent straight lines as shown in the figure 1.

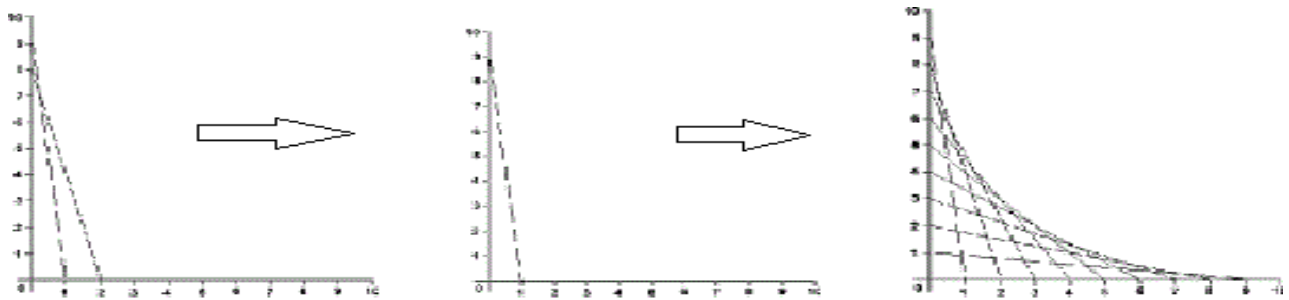

Fig1. (a) curve made up of large number of straight lines-Curve stitching

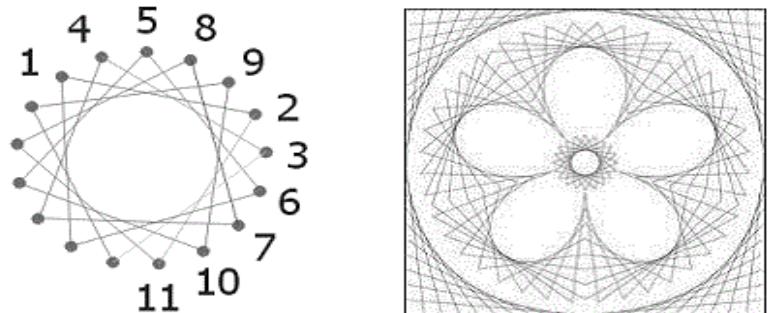

Figure 1.(b) Circle and other complex curves made up of straight line

So taking advantage of this feature, curve detection can be done. Even it can also detect curves having particular range of slopes of tangents lines. I have used Hough lines transform to detect lines which are involved in making the curve. This is done by setting parameters of Hough transform to appropriate values. There are two types of Hough transform:
1) The Standard Hough Transform and
2) The Probabilistic Hough Line Transform.

The Standard Hough Transform returns parameters of detected lines in Polar coordinate a system, which is in a vector of couples $(\theta, r_\theta)$.

The Probabilistic Hough Line Transform more efficient implementation of the Hough Line Transform. It gives as output the extremes of the detected lines $(x_0, y_0, x_1, y_1)$. It is difficult to detect straight lines which are part of a curve because they are very very small. For detecting such lines it is important to properly set all the parameters of Hough transform. Two of most important parameter are: Hough votes and minimum distance between points which are to be joined to make a line. Both of the parameter are set at very less values. Once all the lines which are tangents to the given curve are detected and stored in a vector variable, next step is to find slopes of all these lines by solving equation (1) and (2).

$$y = \left(-\frac{\cos\theta}{\sin\theta}\right)x + \left(\frac{r}{\sin\theta}\right) \qquad (1)$$

$$r_\theta = x_0 \cdot \cos\theta + y_0 \cdot \sin\theta \qquad (2)$$

But when this algorithm was tested, it was found that curves are not detected smoothly and a lot of deviation was present. So for solving this problem, one more algorithm is applied to above algorithm. This algorithm is based on MEAN VALUE THEOREM.

Statement of MEAN VALUE THEOREM: For f: [a, b] → R be a continuous function on the closed interval [a, b], and differentiable on the open interval (a, b), where a < b. Then there exists some c in (a, b) such that

$$f'(c) = \frac{f(b) - f(a)}{b - a}. \qquad (3)$$

$$\lim_{h \to 0} \frac{f(x+h) - f(x)}{h} \qquad (4)$$

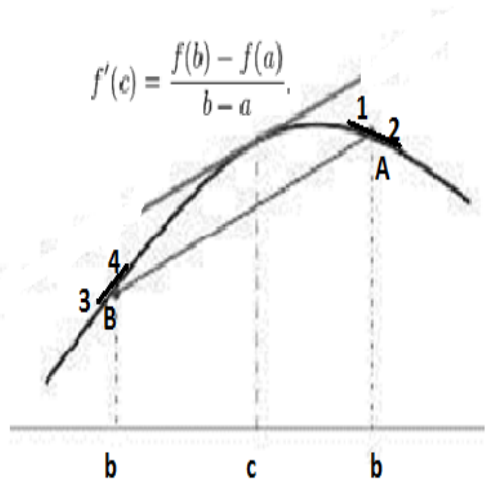

- Points 1, 2, 3, 4 lies on the curve.
- Distance between points 1, 2, 3, 4 is very less.
- 'A' represents Hough line made by joining points '1' and '2'.
- 'B' represents Hough line made by joining points '3' and '4'
- Point 'a' represents centroid of line segment 'A'
- Point 'b' represents centroid of line segment 'B'
- 'c' is point obtained after applying Mean Value Theorem on points 'a' and 'b' which are centroids of line segment A and B.
- Point 'c 'will give a line which is tangent to the curve. Therefore from two Hough lines (which are part of the curve) are used to find one tangent of the curve figure 3. Implementation of mean value theorem on curve using centroids of Hough lines (very small line segments on curve)

In this algorithm, two very small Hough lines are taken on the curve as shown in the figure 3, then weighted centroids of these Hough lines are calculated. These centroids are input to the formula of Mean Value Theorem which returns the slope of tangent at point 'c'. In this way, iterative implementation of above mentioned steps will give many tangents of the curve which provides very efficient tracking of curve in real time as shown in the figure 4(a) and 4(b). The biggest advantage of this algorithm is that it is not dependent on color thresholding techniques because applying color thresholding methods are not robust and may fail in adverse conditions. Even in the case where there are many curves of similar slope characteristics in the same frame, this algorithm can be used to filter out the single desired curve (discussed in the section4). In the section 3, implementation of the above algorithm is shown and explained.

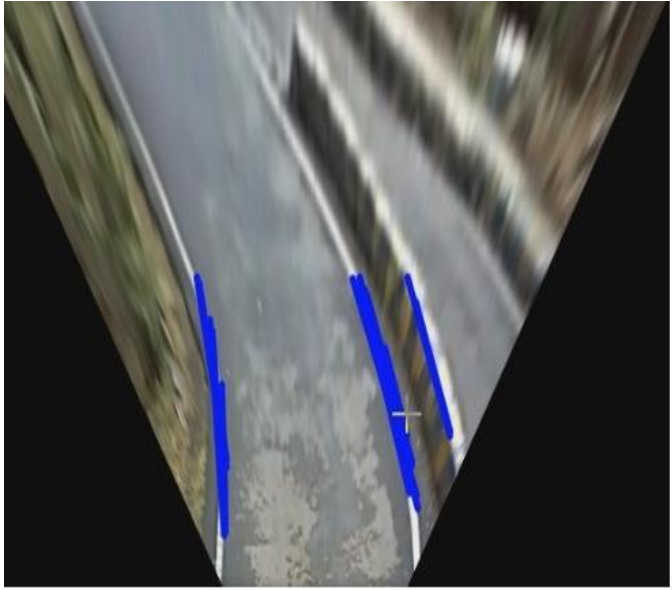

figure 4(a). Birdeye view of detected curvy lanes using mean value theorem on centroids of Hough lines on curve without color thresholding. Noise filter and shadow/illumination correction is also applied here.

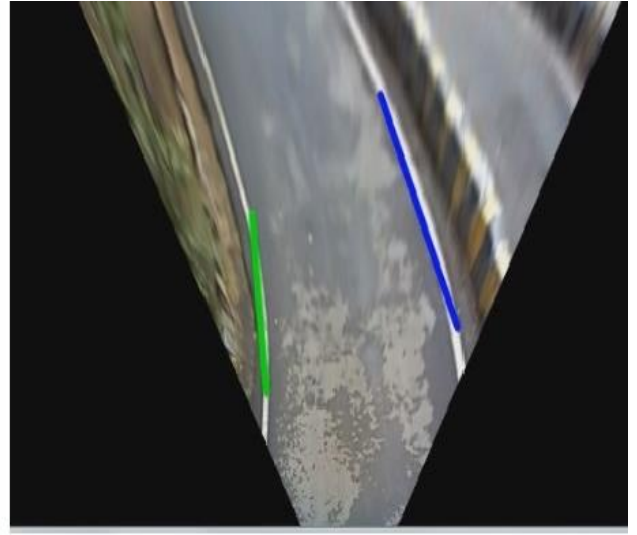

figure 4(b). Birdeye view of detected curvy lanes using mean value theorem on weighted centroids of Hough lines on curve without color thresholding. Noise filter and shadow/illumination correction is also applied here.

## 3. DIFFERENTIATING VARIOUS CURVES USING SLOPES OF HOUGH LINES.

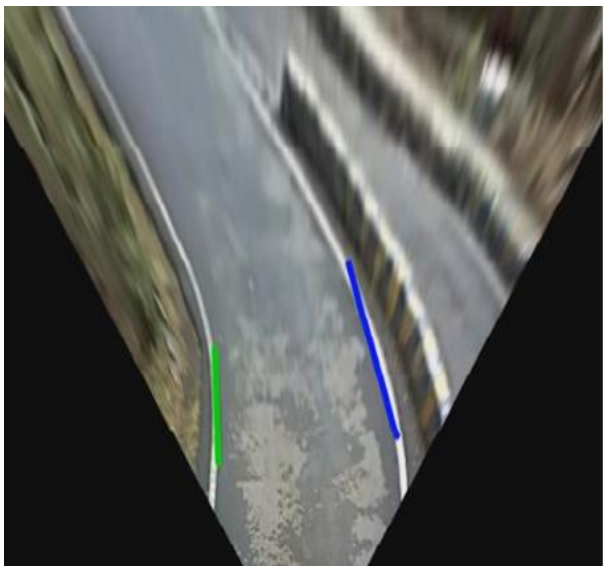

- The figure 5 shows Result of the algorithm for differentiating two lanes by using slopes of the tangents(Obtained by implementing Mean Value Theorem on small Hough lines present on the curves ).
- Blue colored line represents lane which have angle in range: 90 degrees to 120 degrees.
- Green color represents lane which have angle between 60 degrees to 90 degrees.
- The slopes of tangents are calculated using algorithm mentioned in section 1. For finding out slope accurately, distance between the points of a hough line is reduced to very small value, that is limit tending to zero.

$$f'(c) = \frac{f(b) - f(a)}{b - a} \quad \Rightarrow \quad \lim_{h \to 0} \frac{f(x+h) - f(x)}{h}$$

- This algorithm also works well in case of multiple lanes or curves covering 0 to 180 degrees of slopes.

figure 5. Bird eye view of result obtained from algorithm to differentiate various curves using slopes of tangents (from mean value theorem)

For detecting curves of desired slopes, the slopes of the tangents (randomly selected Hough lines on desired curve) of the desired curve are stored in a vector 'A'. During run time detection, the vectors of slopes of tangents of all the curves present in the view are compared with the initially saved vector 'A' (containing slopes of tangents of desired curve). If matching process is 90% successful then desired curve is recognized. The storing and matching of vector takes place in one particular direction (left to right) which means that vector will store slopes of the tangents staring from leftmost point of the curve and similarly matching process starts from the leftmost point of the other curves. So for detecting multiple curves in one frame, more than one vectors can be initialized for matching process and more than one curves can be detected in one frame.

Figures 5(a) given below shows how one particular curve is extracted out from many curves present in single frame by using above method. The slopes of the tangents of the first curve are stored in the vector which is then compared one by one from all the curves and resultant output window will only contain the curve whose slope vector matches with the original vector.

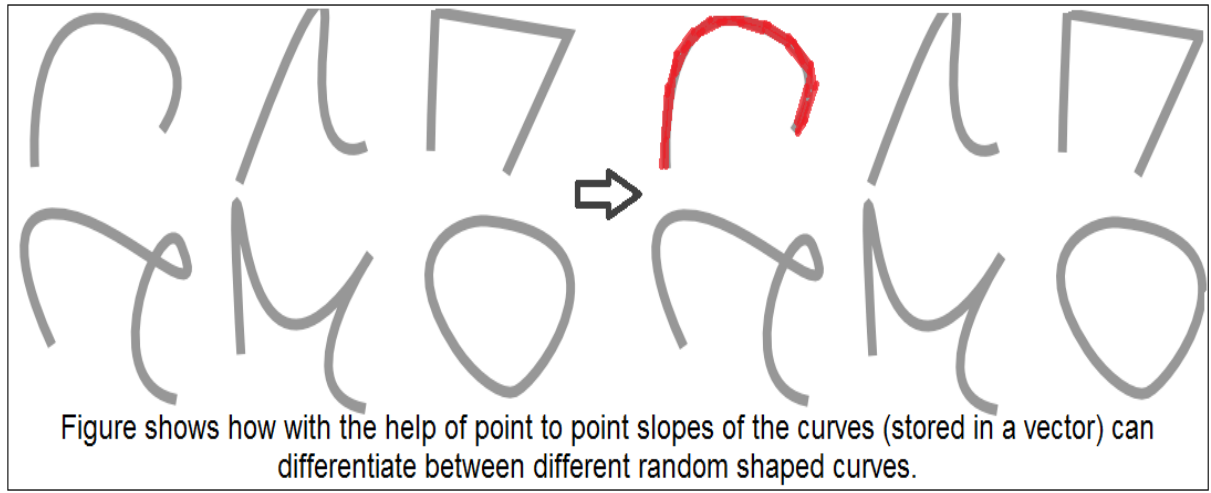

Figure shows how with the help of point to point slopes of the curves (stored in a vector) can differentiate between different random shaped curves.

Figure 5(a)

Figure 5(b) shows how some standard curves like parabola, ellipse, circle and hyperbola can be separately detected by implementing above algorithm.

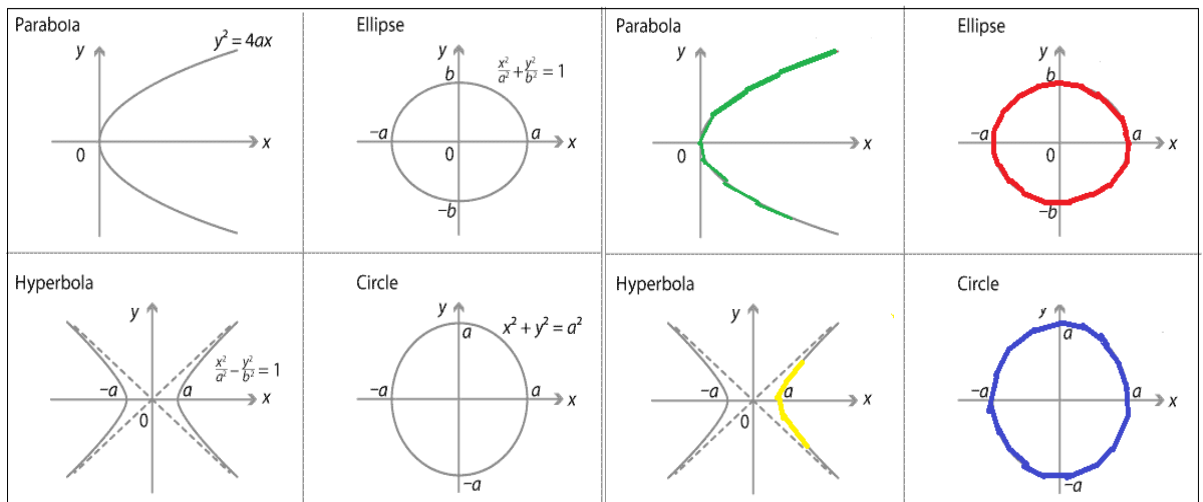

Figure shows how with the help of slope pattern recognition, different standard curves can be seperated out.

Figure 5(b)

So by implementing this algorithm, any type of curve can be detected by storing its slopes (tangents angle) in a vector and then matching it with the subjected curves.

Above given algorithm may also face some problems which may result in incorrect results. Noise and other similar smaller curves are the main reason for wrong results. For example if desired curve to be detected is ellipse, and there are many ellipse present in one frame. In such situation, when algorithm for extracting is applied then the original slope vector of the desired curve may get successfully matched with many ellipse and therefore many such ellipses may be present in the output window. But we want only one ellipse out of so many ellipses. Therefore it is very important to remove these problems to get accurate results. For solving these problems, we used important concepts called clustering of Hough lines, weighted centroid and slope filtering. These concepts and their implementation is described below in section 4. The next section explains the concept of CLUSTERING OF HOUGH LINES USING WEIGHTED CENTROIDS AND ANGLE FILTERING TECHNIQUE.

# 4. Clustering Of Hough Lines Using Weighted Centroids And Angle Filtering Technique.

Apart from desired curves, there may be some other curves with insignificant dimensions and noise present in the region of interest. These small curves and noise may cause serious problem in curve detection and may result in wrong results. To avoid this problem, we used concept of clustering of Hough lines. Clustering of Hough lines means grouping the significant Hough lines together to get more accurate results. This is done by using concept of weighted centroid and angle filtering.

The concept of 'weighted centroid' is different concept then 'centroid' (see in figure 7(a)). 'Weighted Centroid' uses the pixel intensities in the image region as weights in the centroid calculation. With 'Centroid', all pixels in the region get equal weight (see in figure 7(b)).

$$Centroid = \frac{\sum_{n=0}^{N-1} f(n) x(n)}{\sum_{n=0}^{N-1} x(n)}$$

figure 7(a): Weighted centroid, here f(n) is pixel intensity and x(n) is position of corresponding pixel

$$C = \frac{x_1 + x_2 + \cdots + x_k}{k}$$

figure 7(b): Centroid, here x(k) is position of pixel and its intensity is not taken into account.

So if two Hough lines(on curve) are having there weighted centroids very near and there angles are approximately(within certain range) equal then these lines can be replaced by the single line having its weighted centroid at center of individual weighted centroids and its angle equals to average of individual angles. So for detecting desirable curve and remove small curves/noise, a parameter 'vote' is used.

The parameter 'vote' is an integer number representing total number of lines which are having their weighted centroids close enough and their slopes approximately equal. If this parameter 'vote' is above certain threshold value, then the corresponding Hough lines are considered as part of the desirable curve, else all these lines are rejected. The threshold value for the parameter 'vote' can be set according to the condition. It can be seen in figure 8(a) that there are large number of Hough lines on curve (lane) as well as outside the curve (lane). On applying above algorithm, result can be seen in figure 8(b), where only single lines are there corresponding to each curve (lane), which means that all the lines due to noise and small curves were removed.

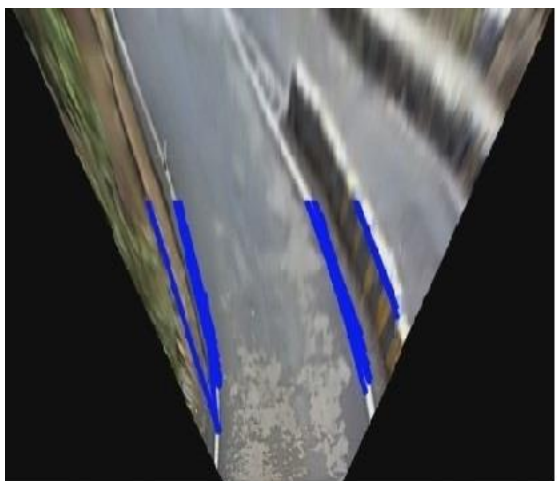

figure 8(a): Bird eye view showing large number of Hough lines on curves (lanes) as well as outside the curve (lane) due to the noise and small curves

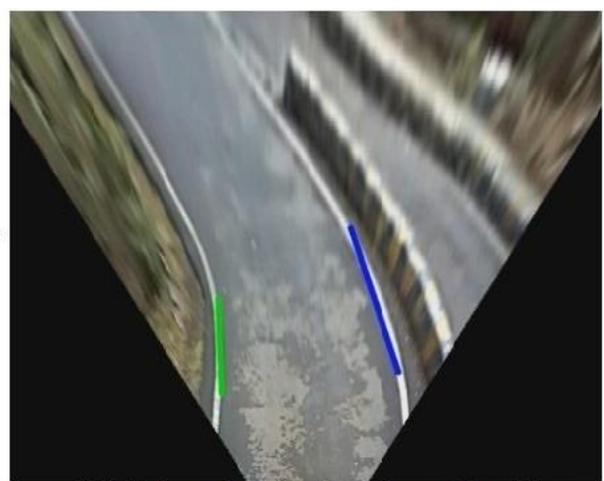

figure 8(b): Bird eye view showing only two Hough lines corresponding to each curve (lane). Problems due to noise and small curves were removed by using weighted centroid and slope filtering algorithms

In the figure given below, there are many ellipses which have same slope (tangent) vectors and thus if algorithm mentioned in section 2 and 3 are applied then all the ellipses may be visible in the output window. Therefore, along with the algorithms mentioned in section 2 and 3, the algorithm mention in section 4 is applied simultaneously. After applying all these algorithm together, only one ellipse is extracted out which have thickest boundaries and all the other small or bigger ellipses are rejected from the output window.

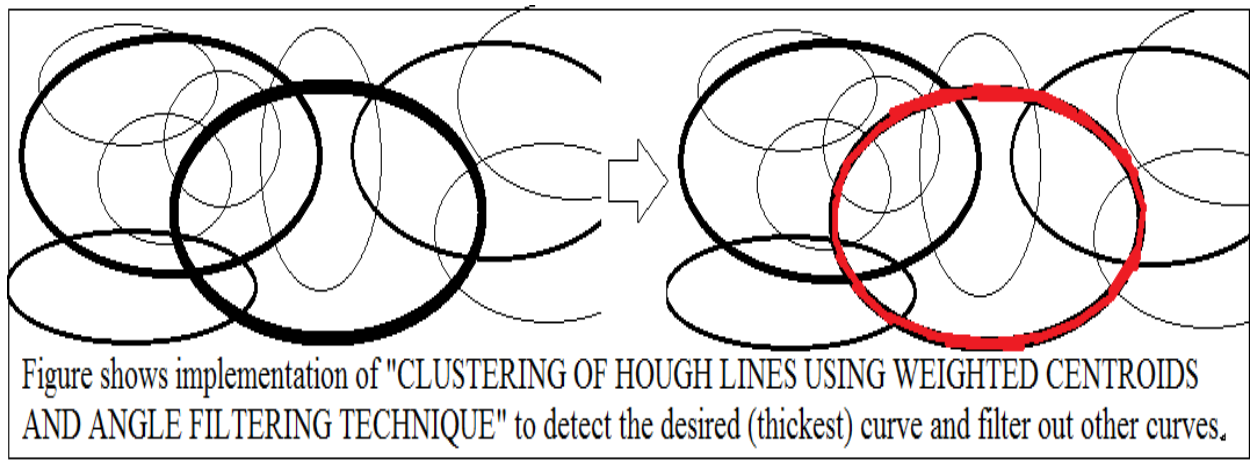

Figure shows implementation of "CLUSTERING OF HOUGH LINES USING WEIGHTED CENTROIDS AND ANGLE FILTERING TECHNIQUE" to detect the desired (thickest) curve and filter out other curves.

## 5. CONTINUOUS FRAME FEEDBACK ALGORITHM FOR DETECTING DISCONTINUOUS CURVES (LANES) AND SHADOW/ILLUMINATION CORRECTION.

In case of discontinuous curve, it is difficult to continue tracking of a curve after even a small discontinuity. For solving this problem, we develop a feedback algorithm which compares the various parameters of previous video frame with current image frame. These are some important parameter which helps to localize the vehicle in between the lanes even in case of discontinuous lanes or absent boundaries. This feedback system is based on three parameter texture analysis, distance analysis and shadow/illumination correction. So the Frame feedback means continuous comparison of current image parameters like texture, distance (perspective vision) with the previous frame of the video feed and fill the discontinuities.

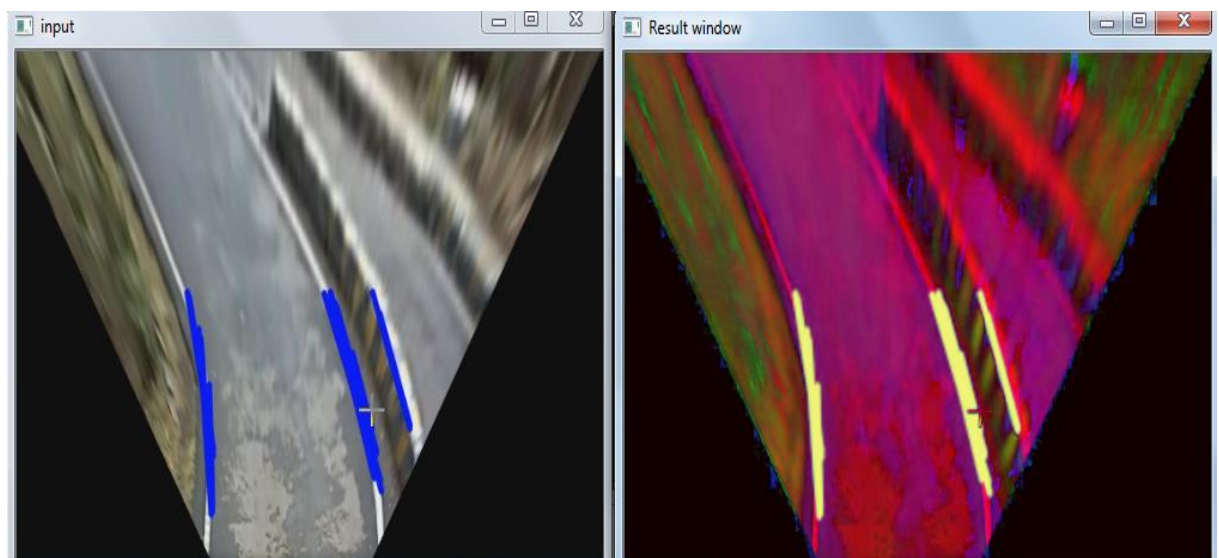

figure 9: In this figure, Result window is showing HSV image frame with noise, shadow and illumination filters. It also includes texture analysis and perspective vision (Bird eye-view).

Analysis of each algorithm: "Shadow/illumination, distance analysis and texture analysis" is discussed one by one below:

DISTANCE ANALYSIS: This is done by using perspective vision, that is by converting normal image frame to bird eye view. This is done for finding out the relation between size of a pixel in a image and real world distance (for example: 10 pixel = 1cm). This helps in localizing the vehicle in between the lanes. If some discontinuity occurs in the lanes, then system automatically maintain the distance according to the parameters obtained from previous image frame. For example in case of curvy lanes, slope of lane can be used by the vehicle to move in particular direction and follow the lanes approximately. But this method is only valid for small discontinuity and if some big discontinuity is present in the lanes then this algorithm may fail. The implementation of this algorithm is shown in the figure 10.

TEXTURE ANALYSIS AND SHADOW/ILLUMINATION CORRECTION: Texture analysis might be applied to various stages of the process. At the preprocessing stage, images was segmented into contiguous regions based on texture properties of each region; At the feature extraction and the classification stages, texture features could provide cues for classifying patterns or identifying objects. As a fundamental basis for all other texture-related applications, texture analysis seeks to derive a general, efficient and compact quantitative description of textures so that various mathematical operations can be used to alter, compare and transform textures. Most available texture analysis algorithms involve extracting texture features and deriving an image coding scheme for presenting selected features.

Figure 10(a) shows the desired curve which is required to be extracted from the live input frame. Figure 10(b) shows the same curve but with many discontinuities. Now the algorithms discussed in previous section will not work properly because of so many discontinuities. Therefore the continuous frame feedback is used to join the discontinuities. So the Frame feedback will continuous compare current image parameters with the previous frame of the video feed and fill the discontinuities.

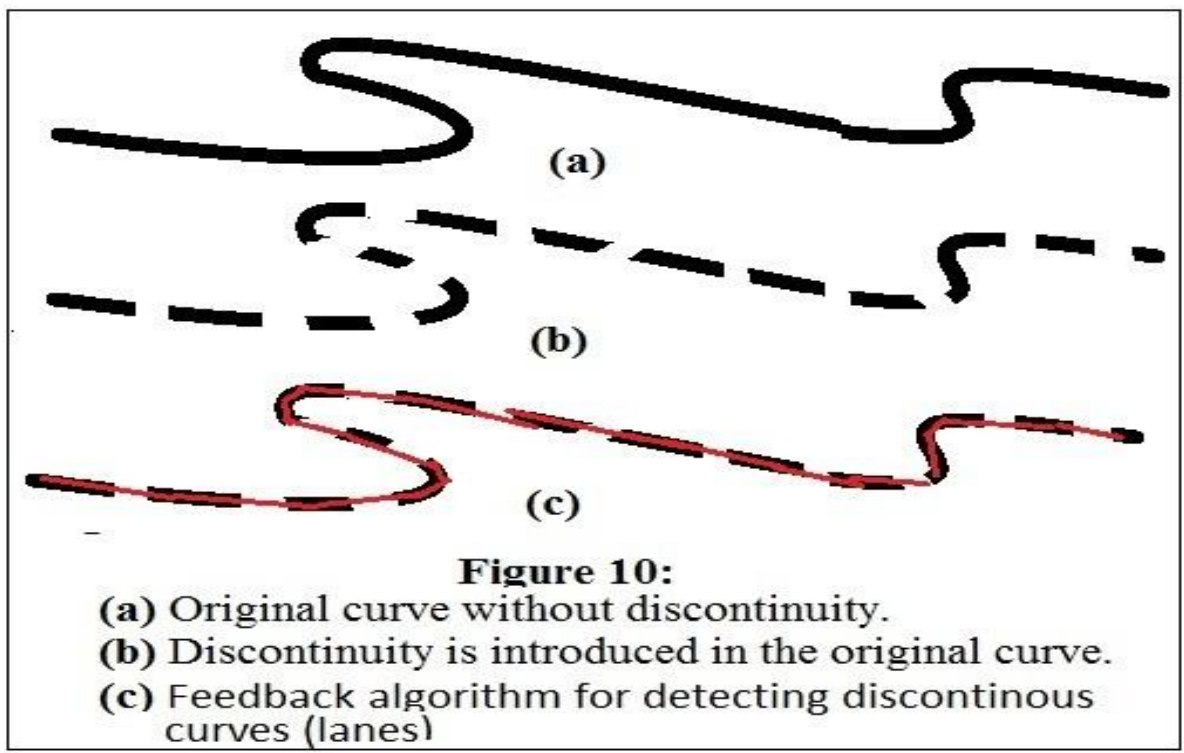

Figure 10:
(a) Original curve without discontinuity.
(b) Discontinuity is introduced in the original curve.
(c) Feedback algorithm for detecting discontinous curves (lanes)

## 7. FLOWCHART

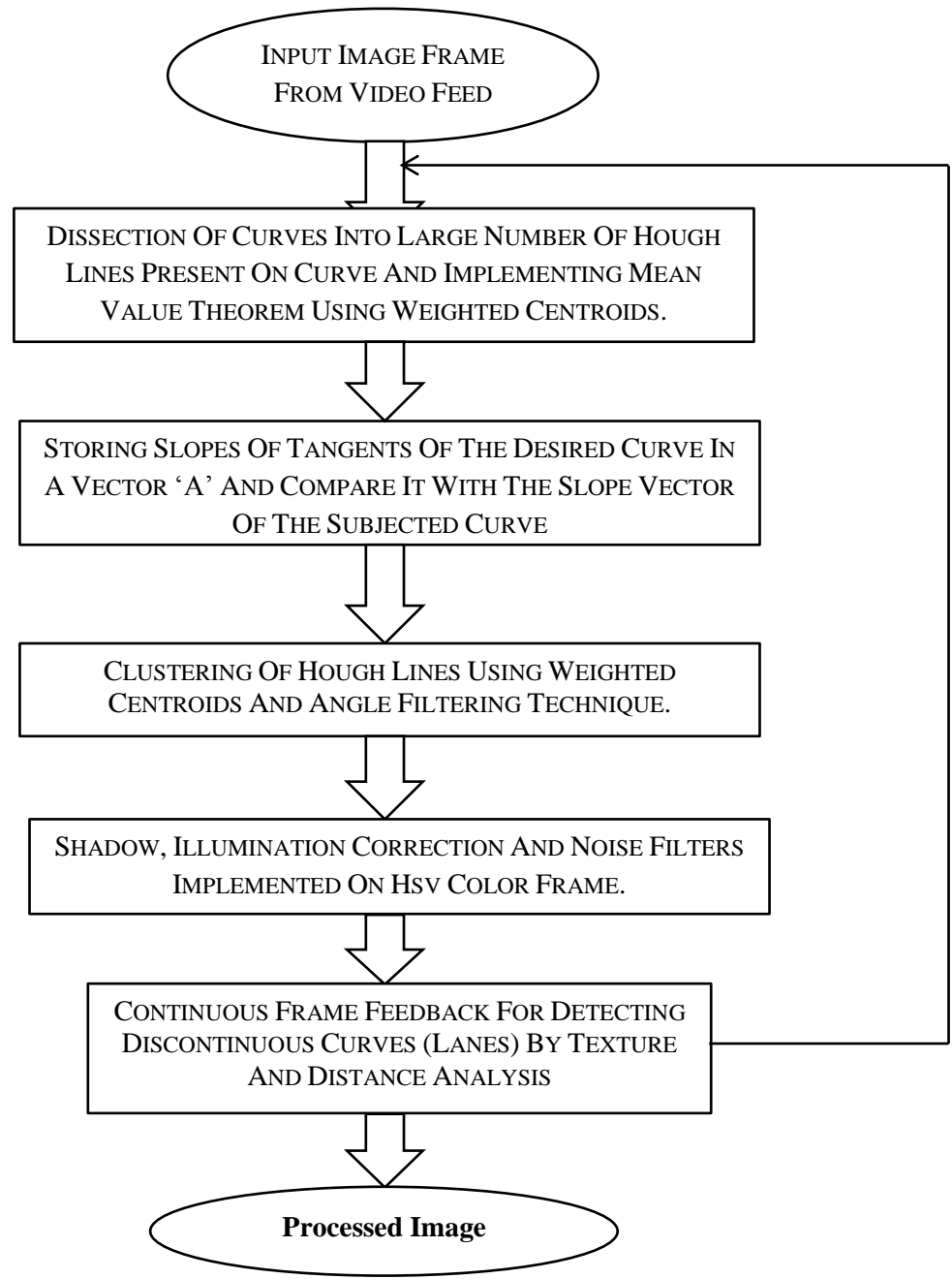

**Flowchart showing combinations of the entire algorithms explained above.**

## 8. CONCLUSION

In this paper we have presented new, unique and robust algorithm for detecting any type of curve (curvy lanes) with the help of tangents of the curve. This algorithm is applicable even in adverse conditions. This algorithm provides a general method to detect any curve with desired slopes. First we dissect the curves into infinite Hough lines (called Curve stitching) and then by applying various algorithm on these Hough lines, we developed a robust algorithm to detect any curvy lanes (or in general: any curves). The implementation of concepts like mean value theorem, clustering of Hough lines, weighted centroids, bird eye view (perspective vision), slope filtering, shadow and illumination correction provides very accurate and efficient algorithm for curvy lane detection or curve detection with desired slopes. I again like to mention that famous libraries like OpenCV, Matlab are only providing functions which are able to detect some standard curves, but this algorithm provides unique feature of detecting any curve having desired range of slopes. Also, this algorithm reduces effect of noise, other small curves, shadow/illumination to great extent. These algorithms can be implemented on autonomous vehicle for robust lane detection and it can also be used in normal vehicle for speed control feedback system to avoid fatal curves on roads. This algorithm can also be used for general purpose curve detection in various image processing applications.

**ACKNOWLEDGEMENT**

I take this opportunity to express my gratitude to the people who have been instrumental in the successful completion of this paper.I am thankful to my institution and innovation cell, iit bombay, for providing me with this opportunity to learn, explore and implement our skills. I also want to thank my family, my seniors Ajinkiya Kamath, Rohan Thakkar, Nikhil Sawake, Aditya Hambarde, Ebrahim Attarwala, Arpit Gupta, Kumar Keshav and my faculties KM Burchandi, Anand Katakwar, Jaiveer Singh and all other special people who directly or indirectly inspired me and motivate me in every step of my life.

## Amartansh Dubey

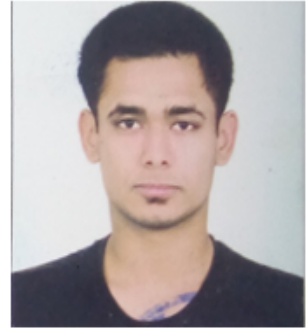

It is evident from past that most of the innovations due to electronics and computer science were never dreamt by common men. I mean no one ever imagine that mobiles,TV,3G,nanomites,sixth sense technology, brain to brain communication,VLSI (few vaccum tubes to millions of transistors on single chip) could ever come to reality and such innovations inspire and motivate me alot, as a result when most of my friends are dreaming about good jobs and salary, i dream about publishing great research papers, becoming part of TED, MS from places like  USA, EUROPE, etc. When i developed autonomous turret(autonomous weapon on wheels), smart wheelchair for paralyzed people, smart sevo motor, FM signal field strength detector, Image processing algorithms for autonomous vehicles,i realised that i with electronics could contribute to my society and may be one day i will become part of some great innovation. Visit my blog for more details: http://electronportal.blogspot.in/

I want to be part of some great research and innovation. Sadly, here in indian society, education is primarily job and salary driven,research always comes second which leaves india far behind in electronics. For motivating my juniors in research, i have started making some very useful tutorials which can be seen on my blog http://electronportal.blogspot.in/. Apart from all these reasons, most important reason for my interest in electronics is level of satisfaction i get in this field. I simply just love electronics and very much passionate about it. Ultimate aim of my life is to become part of some great innovations and research